\documentclass[11pt,a4paper]{article}
\usepackage[nohyperref]{naaclhlt2019}
\usepackage{times}
\usepackage{latexsym}
\usepackage[utf8]{inputenc}

\usepackage{url}

\usepackage{graphicx}
\usepackage{booktabs}
\usepackage{multirow}

\usepackage{amsthm}
\newtheorem{definition}{Definition}
\usepackage{amsmath}
\usepackage{amssymb}
\usepackage{mathtools}

\aclfinalcopy 
\def\aclpaperid{758} 

\title{Simple Question Answering with Subgraph Ranking and Joint-Scoring}

\author{Wenbo Zhao\thanks{\hspace{0.05cm} Work conducted during an internship at Alexa AI, CA.} \\
    Carnegie Mellon University \\
  {\tt wzhao1@andrew.cmu.edu} \\
  \AND
  Tagyoung Chung, Anuj Goyal, \and Angeliki Metallinou \\
    Amazon Alexa AI \\
  {\tt \{tagyoung, anujgoya, ametalli\}@amazon.com} \\}

\date{}

\begin{document}
\maketitle
\begin{abstract}
Knowledge graph based simple question answering (KBSQA) is a major area of research within question answering. Although only dealing with \textit{simple} questions, i.e., questions that can be answered through a single knowledge base (KB) fact, this task is neither simple nor close to being solved. Targeting on the two main steps, subgraph selection and fact selection, the research community has developed sophisticated approaches. However, the importance of subgraph ranking and leveraging the subject--relation dependency of a KB fact have not been sufficiently explored. Motivated by this, we present a unified framework to describe and analyze existing approaches. Using this framework as a starting point, we focus on two aspects: improving subgraph selection through a novel ranking method and leveraging the subject--relation dependency by proposing a joint scoring CNN model with a novel loss function that enforces the well-order of scores. Our methods achieve a new state of the art (85.44\% in accuracy) on the SimpleQuestions dataset.
\end{abstract}

\section{Introduction}
Knowledge graph based simple question answering (KBSQA) is an important area of research within question answering, which is one of the core areas of interest in natural language processing \cite{yao2014information,yih2015semantic,dong2015question,khashabi2016question,zhang2018variational,hu2018answering}. It can be used for many applications such as virtual home assistants, customer service, and chat-bots. A {\it knowledge graph} is a multi-entity and multi-relation directed graph containing the information needed to answer the questions. The graph can be represented as collection of triples \{(subject, relation, object)\}. Each triple is called a {\it fact}, where a directed relational arrow points from subject node to object node. A {\it simple} question means that the question can be answered by extracting a single fact from the knowledge graph, i.e., the question has a single subject and a single relation, hence a single answer. For example, the question ``Which Harry Potter series did Rufus Scrimgeour appear in?'' can be answered by a single fact (Rufus Scrimgeour, book.book-characters.appears-in-book, Harry Potter and the Deathly Hallows). Given the simplicity of the questions, one would think this task is trivial. Yet it is far from being easy or close to being solved. The complexity lies in two aspects. One is the massive size of the knowledge graph, usually in the order of billions of facts. The other is the variability of the questions in natural language. Based on this anatomy of the problem, the solutions also consist of two steps: (1) selecting a relatively small subgraph from the knowledge graph given a question and (2) selecting the correct fact from the subgraph.

Different approaches have been studied to tackle the KBSQA problems. The common solution for the first step, subgraph selection (which is also known as entity linking), is to label the question with subject part (\emph{mention}) and non-subject part (\emph{pattern}) and then use the mention to retrieve related facts from the knowledge graph, constituting the subgraph. Sequence labeling models, such as a BiLSTM-CRF tagger~\cite{huang2015bidirectional}, are commonly employed to label the mention and the pattern. To retrieve the subgraph, it is common to search all possible $n$-grams of the mention against the knowledge graph and collect the facts with matched subjects as the subgraph. The candidate facts in the subgraph may contain incorrect subjects and relations. In our running example, we first identify the mention in the question, i.e.,``Rufus Scrimgeour'', and then retrieve the subgraph which could contain the following facts: \{(Rufus Scrimgeour, book.book-characters.appears-in-book, Harry Potter and the Deathly Hallows), (Rufus Wainwright, music.singer.singer-of, I Don't Know What That Is)\}.

For the second step, fact selection, a common approach is to construct models to match the mention with candidate subjects and match the pattern with candidate relations in the subgraph from the first step. For example, the correct fact is identified by matching the mention ``Rufus Scrimgeour'' with candidate subjects \{Rufus Scrimgeour, Rufus Wainwright\} and matching the pattern ``Which Harry Potter series did $m$ appear in'' with candidate relations \{book.book-characters.appears-in-book, music.singer.singer-of\}. Different neural network models can be employed~\cite{bordes2015large,dai-li-xu:2016:P16-1,yin-EtAl:2016:COLING,yu2017improved,petrochuk2018simplequestions}.

Effective as these existing approaches are, there are three major drawbacks. (1) First, in subgraph selection, there is no effective way to deal with inexact matches and the facts in subgraph are not ranked by relevance to the mention; however, we will later show that effective ranking can substantially improve the subgraph recall.
(2) Second, the existing approaches do not leverage the dependency between mention--subjects and pattern--relations; however, mismatches of mention--subject can lead to incorrect relations and hence incorrect answers. We will later show that leveraging such dependency contributes to the overall accuracy.
(3) Third, the existing approaches minimize the ranking loss
~\cite{yin-EtAl:2016:COLING,lukovnikov2017neural,qu2018question}; however, we will later show that the ranking loss is suboptimal.

Addressing these points, the contributions of this paper are three-fold:
(1) We propose a subgraph ranking method with combined literal and semantic score to improve the recall of the subgraph selection. It can deal with inexact match, and achieves better performance compared to the previous state of the art.
(2) We propose a low-complexity joint-scoring CNN model and a well-order loss to improve fact selection. It couples the subject matching and the relation matching by learning order-preserving scores and dynamically adjusting the weights of scores.
(3) We achieve better performance (85.44\% in accuracy) than the previous state of the art on the SimpleQuestions dataset, surpassing the best baseline by a large margin\footnote{\citet{ture-jojic:2017:EMNLP2017} reported better performance than us but neither \citet{petrochuk2018simplequestions} nor \citet{mohammed2017strong} could replicate their result.}.

\section{Related Work}
The methods for subgraph selection fall in two schools: parsing methods~\cite{berant2013semantic,yih2015semantic,zheng2018question} and sequence tagging methods~\cite{yin-EtAl:2016:COLING}. The latter proves to be simpler yet effective, with the most effective model being BiLSTM-CRF~\cite{yin-EtAl:2016:COLING,dai-li-xu:2016:P16-1,petrochuk2018simplequestions}.

The two categories of methods for fact selection are match-scoring models and classification models. The match-scoring models employ neural networks to score the similarity between the question and the candidate facts in the subgraph and then find the best match. For instance, \citet{bordes2015large} use a memory network to encode the questions and the facts to the same representation space and score their similarities. \citet{yin-EtAl:2016:COLING} use two independent models, a character-level CNN and a word-level CNN with attentive max-pooling. \citet{dai-li-xu:2016:P16-1} formulate a two-step conditional probability estimation problem and use BiGRU networks.
\citet{yu2017improved} use two separate hierarchical residual BiLSTMs to represent questions and relations at different abstractions and granularities. \citet{qu2018question} propose an attentive recurrent neural network with similarity matrix based convolutional neural network (AR-SMCNN) to capture the semantic-level and literal-level similarities.
In the classification models, \citet{ture-jojic:2017:EMNLP2017} employ a two-layer BiGRU model. \citet{petrochuk2018simplequestions} employ a BiLSTM to classify the relations and achieve the state-of-the-art performance. In addition, \citet{mohammed2017strong} evaluate various strong baselines with simple neural networks (LSTMs and GRUs) or non-neural network models (CRF). \citet{lukovnikov2017neural} propose an end-to-end word/character-level encoding network to rank subject--relation pairs and retrieve relevant facts.

However, the multitude of methods yield progressively smaller gains with increasing model complexity~\cite{mohammed2017strong,gupta2018retrieve}. Most approaches focus on fact matching and relation classification while assigning less emphasis to subgraph selection. They also do not sufficiently leverage the important signature of the knowledge graph---the subject--relation dependency, namely, incorrect subject matching can lead to incorrect relations. Our approach is similar to~\cite{yin-EtAl:2016:COLING}, but we take a different path by focusing on accurate subgraph selection and utilizing the subject--relation dependency.

\section{Question Answering with Subgraph Ranking and Joint-Scoring}
\subsection{Unified Framework}
We provide a unified description of the KBSQA framework. First, we define
\begin{definition}
\textup{\textbf{Answerable Question}}
A question is answerable if and only if one of its facts is in the knowledge graph.
\end{definition}

\noindent Let $\mathcal{Q} \coloneqq \{q \mid q \text{ is anwerable}\}$ be the set of answerable questions, and $\mathcal{G} \coloneqq \{(s, r, o) \mid s \in \mathcal{S}, r \in \mathcal{R}, o \in \mathcal{O}\}$ be the knowledge graph, where $\mathcal{S}$, $\mathcal{R}$ and $\mathcal{O}$ are the set of subjects, relations and objects, respectively. The triple $(s, r, o)$ is a \emph{fact}.
By the definition of answerable questions, the key to solving the KBSQA problem is to \emph{find the fact in knowledge graph corresponding to the question}, i.e., we want a map $\Phi: \mathcal{Q} \to \mathcal{G}$.
Ideally, we would like this map to be injective such that for each question, the corresponding fact can be uniquely determined (more precisely, the injection maps from the {\it equivalent class} of $\mathcal{Q}$ to $\mathcal{G}$ since similar questions may have the same answer, but we neglect such difference here for simplicity). However, in general, it is hard to find such map directly because of (1) the massive knowledge graph and (2) natural language variations in questions. Therefore, end-to-end approaches such as parsing to structured query and encoding-decoding models are difficult to achieve~\cite{yih2015semantic,sukhbaatar2015end,kumar2016ask,he2016character,hao2017end}.
Instead, related works and this work mitigate the difficulties by breaking down the problem into the aforementioned two steps, as illustrated below:

{\small
\setlength{\abovedisplayskip}{0pt}
\setlength{\belowdisplayskip}{7pt}
\setlength{\abovedisplayshortskip}{0pt}
\setlength{\belowdisplayshortskip}{0pt}
\begin{align*}
&\textbf{(1) Subgraph Selection:} \\
&\text{question} \longrightarrow \{\text{mention, pattern}\},\quad \text{mention} \longrightarrow \text{subgraph} \\ \\
&\textbf{(2) Fact Selection:} \\
&\text{match}
\begin{cases}
	\text{mention} \leftrightarrow \text{subject} \\
	\text{pattern} \leftrightarrow \text{relation}
\end{cases}
\forall (\text{subject, relation}) \in \text{subgraph} \\
&\Rightarrow \text{(subject*, relation*)} \longrightarrow \text{object* (answer*)}
\end{align*}
}%
In the first step, the size of the knowledge graph is significantly reduced. In the second step, the variations of questions are confined to mention--subject variation and pattern--relation variation.

Formally, we denote the questions as the union of mentions and patterns $\mathcal{Q} = \mathcal{M}\bigcup\mathcal{P}$ and the knowledge graph as the subset of the Cartesian product of subjects, relations and objects $\mathcal{G} \subseteq \mathcal{S}\times\mathcal{R}\times\mathcal{O}$.
In the first step, given a question $q \in \mathcal{Q}$, we find the mention via a sequence tagger $\tau: \mathcal{Q} \to \mathcal{M}$, $q \mapsto m_q$. The tagged mention consists of a sequence of words $m_q = \{w_1, \ldots, w_n\}$ and the pattern is the question excluding the mention $p_q = q \backslash m_q$. We denote the set of $n$-grams of $m_q$ as $\mathcal{W}_n(m_q)$ and use $\mathcal{W}_n(m_q)$ to retrieve the subgraph as $\mathcal{S}_q \times \mathcal{R}_q \times \mathcal{O}_q \supseteq \mathcal{G}_q \coloneqq \{(s, r, o) \in \mathcal{G} \mid \mathcal{W}_n(s) \bigcap \mathcal{W}_n(m_q) \neq \varnothing, n = 1, \ldots, |m_q| \}$.

Next, to select the correct fact (the answer) in the subgraph, we match the mention $m_q$ with candidate subjects in $\mathcal{S}_q$, and match the pattern $p_q$ with candidate relations in $\mathcal{R}_q$. Specifically, we want to maximize the log-likelihood

{\small
\setlength{\abovedisplayskip}{0pt}
\setlength{\belowdisplayskip}{7pt}
\setlength{\abovedisplayshortskip}{0pt}
\setlength{\belowdisplayshortskip}{0pt}
\begin{align}
\begin{cases}
 \max_{s \in \mathcal{S}_q} \log\mathbb{P}(s \mid m_q) \\
 \max_{r \in \mathcal{R}_q} \log\mathbb{P}(r \mid p_q).
\end{cases}
\label{eq:match_max_prob}
\end{align}
}%
The probabilities in (\ref{eq:match_max_prob}) are modeled by

{\small
\setlength{\abovedisplayskip}{0pt}
\setlength{\belowdisplayskip}{7pt}
\setlength{\abovedisplayshortskip}{0pt}
\setlength{\belowdisplayshortskip}{0pt}
\begin{align}
    \mathbb{P}(s \mid m_q) = \frac{e^{h(f(m_q), f(s))}}{\sum_{s' \in \mathcal{S}_q} e^{h(f(m_q), f(s'))}} \label{eq:prob_model_1}\\
    \mathbb{P}(r \mid p_q) = \frac{e^{h(g(p_q), g(r))}}{\sum_{r' \in \mathcal{R}_q} e^{h(g(p_q), g(r'))}} \label{eq:prob_model_2},
\end{align}
}%
where $f: \mathcal{M} \bigcup \mathcal{S} \to \mathbb{R}^d$ maps the mention and the subject onto a $d$-dimensional differentiable manifold embedded in the Hilbert space and similarly, $g: \mathcal{P} \bigcup \mathcal{R} \to \mathbb{R}^{d}$. Both $f$ and $g$ are in the form of neural networks. The map $h: \mathbb{R}^d \times \mathbb{R}^d \to \mathbb{R}$ is a metric that measures the similarity of the vector representations  (e.g., the cosine similarity).
Practically, directly optimizing (\ref{eq:match_max_prob}) is difficult because the subgraph $\mathcal{G}_q$ is large and computing the partition functions in (\ref{eq:prob_model_1}) and (\ref{eq:prob_model_2}) can be intractable.
Alternatively, a surrogate objective, the ranking loss (or hinge loss with negative samples)~\cite{collobert2008unified,dai-li-xu:2016:P16-1} is minimized

{\small
\setlength{\abovedisplayskip}{0pt}
\setlength{\belowdisplayskip}{7pt}
\setlength{\abovedisplayshortskip}{0pt}
\setlength{\belowdisplayshortskip}{0pt}
\begin{align}
\mathcal{L}_{\text{rank}} & = 
\sum_{q \in \mathcal{Q}} \left( \sum_{s \in \mathcal{S}_q} \left[ h_{f}(m_q, s^{-}) - h_{f}(m_q, s^{+}) + \lambda \right]_+ \right. \nonumber \\
& + \left. \sum_{r \in \mathcal{R}_q} \left[ h_{g}(p_q, r^{-}) - h_g(p_q, r^{+}) + \lambda \right]_+ \right),
\label{eq:rank_loss}
\end{align}
}%
where $h_{f}(\cdot, \cdot) = h(f(\cdot), f(\cdot))$, $h_{g}(\cdot, \cdot) = h(g(\cdot), g(\cdot))$; the sign $+$ and $-$ indicate correct candidate and incorrect candidate, $[\cdot]_+ = \max(\cdot, 0)$, and $\lambda > 0$ is a margin term. Other variants of the ranking loss are also studied~\cite{cao2006adapting,zhao2015deep,vu2016bi}.

\subsection{Subgraph Ranking}
\label{ssec:subgraph_rank}
To retrieve the subgraph of candidate facts using $n$-gram matching~\cite{bordes2015large}, one first constructs the map from $n$-grams $\mathcal{W}_n(s)$ to subject $s$ for all subjects in the knowledge graph, yielding $\{\mathcal{W}_n(s) \to s \mid s \in \mathcal{S}, n = 1, \ldots, |s|\}$. Next, 
one uses the $n$-grams of mention $\mathcal{W}_n(m)$ to match the $n$-grams of subjects $\mathcal{W}_n(s)$ and fetches those matched facts to compose the subgraph $\{(s, r, o) \in \mathcal{G} \mid \mathcal{W}_n(s) \bigcap \mathcal{W}_n(m) \neq \varnothing, n = 1, \ldots |m|\}$. In our running example, for the mention ``Rufus Scrimgeour'', we collect the subgraph of facts with the bigrams and unigrams of subjects matching the bigram \{``Rufus Scrimgeour''\} and unigrams \{``Rufus'', ``Scrimgeour''\}.

One problem with this approach is that the retrieved subgraph can be fairly large. Therefore, it is desirable to rank the subgraph by relevance to the mention and only preserve the most relevant facts. To this end, different ranking methods are used, such as surface-level matching score with added heuristics~\cite{yin-EtAl:2016:COLING}, relation detection network~\cite{yu2017improved,hao2018pattern}, term frequency-inverse document frequency (TF-IDF) score~\cite{ture-jojic:2017:EMNLP2017,mohammed2017strong}. However, these ranking methods only consider matching surface forms and cannot handle inexact matches, synonyms, or polysemy (``New York'' , ``the New York City'', ``Big Apple'').

This motivates us to rank the subgraph not only by literal relevance but also semantic relevance. Hence, we propose a ranking score with literal closeness and semantic closeness. Specifically, the literal closeness is measured by the length of the longest common subsequence $|\sigma|(s, m)$ between a subject $s$ and a mention $m$. The semantic closeness is measured by the co-occurrence probability of the subject $s$ and the mention $m$

{\small
\setlength{\abovedisplayskip}{0pt}
\setlength{\belowdisplayskip}{7pt}
\setlength{\abovedisplayshortskip}{0pt}
\setlength{\belowdisplayshortskip}{0pt}
\begin{align}
\mathbb{P}(s, m) &= \mathbb{P}(s | m) \mathbb{P}(m) \nonumber \\
&=\mathbb{P}(w_1, \ldots w_n | \widetilde{w}_1, \ldots \widetilde{w}_m)\mathbb{P}(\widetilde{w}_1, \ldots \widetilde{w}_m) \label{eq:prob_2}\\
&= \prod_{i=1}^{n}\mathbb{P}(w_i | \widetilde{w}_1, \ldots \widetilde{w}_m)\mathbb{P}(\widetilde{w}_1, \ldots \widetilde{w}_m) \label{eq:prob_3}\\
&=  \prod_{i=1}^{n}\left(\prod_{k=1}^{m}\mathbb{P}(w_i | \widetilde{w}_k)\right) \mathbb{P}(\widetilde{w}_1, \ldots \widetilde{w}_m) \label{eq:prob_4}\\
&= \prod_{i=1}^{n}\left(\prod_{k=1}^{m}\mathbb{P}(w_i | \widetilde{w}_k)\right) \prod_{j=1}^{m-1} \mathbb{P}(\widetilde{w}_{j+1} | \widetilde{w}_{j})\mathbb{P}(\widetilde{w}_1), \label{eq:prob_5}
\end{align}
}%
where from (\ref{eq:prob_2}) to (\ref{eq:prob_3}) we assume conditional independence of the words in subject and the words in mention; from (\ref{eq:prob_3}) to (\ref{eq:prob_4}) and from (\ref{eq:prob_4}) to (\ref{eq:prob_5}) we factorize the factors using the chain rule with conditional independence assumption.
The marginal term $\mathbb{P}(\widetilde{w}_1)$ is calculated by the word occurrence frequency. 
Each conditional term is approximated by $\mathbb{P}(w_i | w_j) \approx \exp\{ \hat{w}_i^T \hat{w}_j\}$ where $\hat{w}_i$s are pretrained GloVe vectors~\cite{pennington-socher-manning:2014:EMNLP2014}. These vectors are obtained by taking into account the word co-occurrence probability of surrounding context. Hence, the GloVe vector space encodes the semantic closeness.
In practice we use the log-likelihood as the semantic score to convert multiplication in (\ref{eq:prob_5}) to summation and normalize the GloVe embeddings into a unit ball.
Then, the score for ranking the subgraph is the weighted sum of the literal score and the semantic score

{\small
\setlength{\abovedisplayskip}{0pt}
\setlength{\belowdisplayskip}{7pt}
\setlength{\abovedisplayshortskip}{0pt}
\setlength{\belowdisplayshortskip}{0pt}
\begin{align}
\text{score}(s, m) = \tau |\sigma|(s, m) + (1 - \tau)\log{\mathbb{P}(s, m)},
\label{eq:relevance_score}
\end{align}
}%
where $\tau$ is a hyper-parameter whose value need to be tuned on the validation set. Consequently, for each question $q$, we can get the top-$n$ ranked subgraph $\mathcal{G}_{q\downarrow}^n$ as well as the corresponding top-$n$ ranked candidate subjects $\mathcal{S}_{q\downarrow}^n$ and relations $\mathcal{R}_{q\downarrow}^n$.

\subsection{Joint-Scoring Model with Well-Order Loss}
\label{ssec:js_wellorder}
\begin{figure*}[t!]
\centering
  \includegraphics[width=0.85\linewidth]{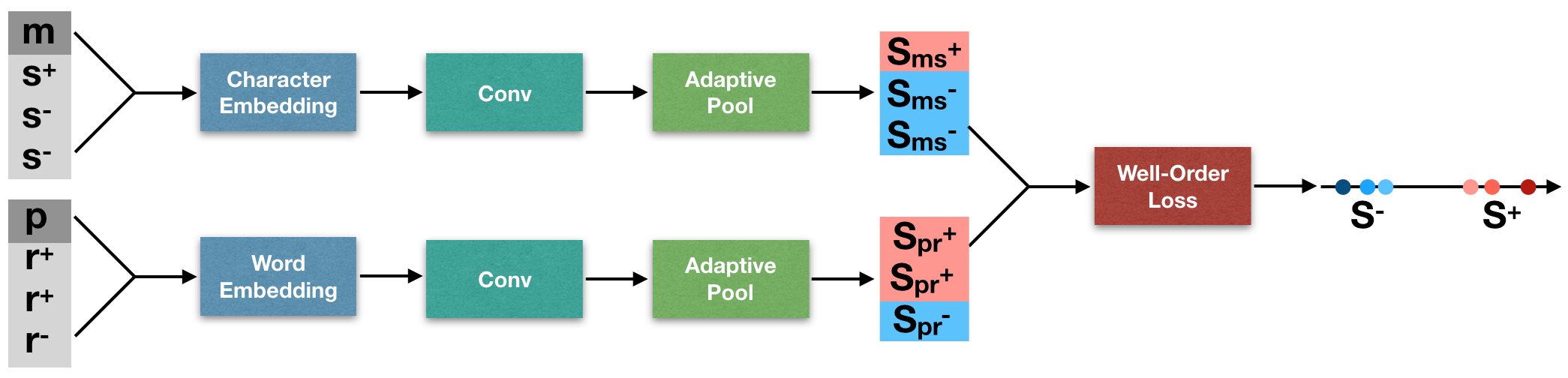}
  \caption{Model Diagram (Section~\ref{ssec:js_wellorder}) {The model takes input pairs (\textbf{m}ention, \textbf{s}ubject) and (\textbf{p}attern, \textbf{r}elation) to produce the similarity scores. The loss dynamically adjusts the weights and enforces the order of positive and negative scores.}}
  \label{fig:js_model}
\end{figure*}

Once we have the ranked subgraph, next we need to identify the correct fact in the subgraph. One school of conventional methods~\cite{bordes-chopra-weston:2014:EMNLP2014,bordes2015large,yin-EtAl:2016:COLING,dai-li-xu:2016:P16-1} is minimizing the surrogate ranking loss (\ref{eq:rank_loss}) where neural networks are used to transform the (subject, mention) and (relation, pattern) pairs into a Hilbert space and score them with inner product.

One problem with this approach is that it matches mention--subject and pattern--relation separately, neglecting the difference of their contributions to fact matching. Given that the number of subjects (order of millions) are much larger than the number of relations (order of thousands), incorrect subject matching can lead to larger error than incorrect relation matching. Therefore, matching the subjects correctly should be given more importance than matching the relations.
Further, the ranking loss is suboptimal, as it does not preserve the relative order of the matching scores. We empirically find that the ranking loss tends to bring the matching scores to the neighborhood of zero (during the training the scores shrink to very small numbers), which is not functioning as intended.

To address these points, we propose a joint-scoring model with well-order loss (Figure~\ref{fig:js_model}). Together they learn to map from joint-input pairs to order-preserving scores supervised by a well-order loss, hence the name.
The joint-scoring model takes joint-input pairs, (subject, mention) or (relation, pattern), to produce the similarity scores directly.
The well-order loss then enforces the well-order in scores.

A well-order, first of all, is a total order---a binary relation on a set which is antisymmetric, transitive, and connex. In our case it is just the ``$\leq$'' relation. In addition, the well-order is a total order with the property that every non-empty set has a least element. 
The well-order restricts that the scores of correct matches are always larger or equal to the scores of incorrect matches, i.e., $\forall i: \forall j: S_i^+ \geq S_j^-$ where $S_i^+$ and $S_i^-$ indicate the score of correct match and the score of incorrect match.

We derive the well-order loss in the following way. Let $S = \{S_1, \ldots, S_n\} = S^+ \bigcup S^-$ be the set of scores where $S^+$ and $S^-$ are the set of scores with correct and incorrect matches. Let $I = I^+ \bigcup I^-$ be the index set of $S$, $|I^+| = n_1$, $|I^-| = n_2$, $n = n_1 + n_2$. Following the well-order relation

{\small
\setlength{\abovedisplayskip}{0pt}
\setlength{\belowdisplayskip}{0pt}
\setlength{\abovedisplayshortskip}{0pt}
\setlength{\belowdisplayshortskip}{0pt}
\begin{align}
& \inf{S^+}\geq \sup{S^-} \nonumber \\
\Leftrightarrow  &\; \forall i^+ \in I^+: \forall i^- \in I^-: S_{i^+}^+ - S_{i^-}^- \geq 0 \nonumber\\ 
\Leftrightarrow &\; \sum_{i^+ \in I^+} \sum_{i^- \in I^-} (S_{i^+}^+ - S_{i^-}^-) \geq 0 \label{eq:bef} \\
\Leftrightarrow &\; n_2 \sum_{i^+ \in I^+} S_{i^+}^+ - n_1 \sum_{i^- \in I^-} S_{i^-}^- \geq 0, \label{eq:aft} \\
\nonumber
\end{align}
}%
where from (\ref{eq:bef}) to (\ref{eq:aft}) we expand the sums and reorder the terms. Consequently, we obtain the well-order loss

{\small
\setlength{\abovedisplayskip}{0pt}
\setlength{\belowdisplayskip}{7pt}
\setlength{\abovedisplayshortskip}{0pt}
\setlength{\belowdisplayshortskip}{0pt}
\begin{align}
&\mathcal{L}_{\text{well-order}}(S_{ms}, S_{pr}) = \nonumber \\
&\left[|I^+| \sum_{i^-} S_{ms}^{i^-} - |I^-| \sum_{i^+} S_{ms}^{i^+} + |I^+||I^-|\lambda\right]_+ + \nonumber \\
&\left[|J^+| \sum_{j^-} S_{pr}^{j^-} - |J^-| \sum_{j^+} S_{pr}^{j^+} + |J^+||J^-|\lambda\right]_+,
\label{eq:well_order_loss}
\end{align}
}%
where  $S_{ms}$, $S_{pr}$ are the scores for (mention, subject), (pattern, relation) pairs for a question, $I$, $J$ are the index sets for candidate subjects, relations in the ranked subgraph, $+$, $-$ indicate the correct candidate and incorrect candidate, $[\cdot]_+ = \max(\cdot, 0)$, and $\lambda > 0$ is a margin term.
Then, the objective (\ref{eq:match_max_prob}) becomes

{\small
\setlength{\abovedisplayskip}{0pt}
\setlength{\belowdisplayskip}{7pt}
\setlength{\abovedisplayshortskip}{0pt}
\setlength{\belowdisplayshortskip}{0pt}
\begin{align}
	\min_{q \in \mathcal{Q}, (s, r) \in \mathcal{S}_{q\downarrow}^n \times \mathcal{R}_{q\downarrow}^n} 
		&\left[ |I^+| \sum_{i^-} h_f(m_q, s^{i^-}) - \right. \nonumber\\
		&\quad  \left. |I^-| \sum_{i^+} h_f(m_q, s^{i^+}) + |I^+| |I^-| \lambda \right]_+ \nonumber\\
	   +&\left[ |J^+| \sum_{j^-} h_g(p_q, r^{j^-}) - \right. \nonumber\\
		&\quad  \left. |J^-| \sum_{j^+} h_g(p_q, r^{j^+}) + |J^+| |J^-| \lambda \right]_+.
	\label{eq:objective_wellorder}
\end{align}
}%
This new objective with well-order loss differs from the ranking loss (\ref{eq:rank_loss}) in two ways, and plays a vital role in the optimization. First, 
instead of considering the match of mention--subjects and pattern--relations separately, (\ref{eq:objective_wellorder}) {\it jointly considers both input pairs and their dependency}. Specifically, (\ref{eq:objective_wellorder}) incorporates such dependency as the weight factors $|I|$ (for subjects) and $|J|$ (for relations). These factors are the controlling factors and are automatically and dynamically adjusted as they are the sizes of candidate subjects and relations.
Further, the match of subjects, weighted by ($I^+$, $I^-$), will control the match of relations, weighted by ($J^+$, $J^-$). To see this, for a question and a fixed number of candidate facts in subgraph, $|I| = |J|$, the incorrect number of subjects $|I^-|$ is usually larger than the incorrect number of relations $|J^-|$, which causes larger loss for mismatching subjects. As a result, the model is forced to match subjects more correctly, and in turn, prune the relations with incorrect subjects and reduce the size of $J^-$, leading to smaller loss. Second, the well-order loss enforces the well-order relation of scores while the ranking loss does not have such constraint.

\section{Experiments}
Here, we evaluate our proposed approach for the KBSQA problem on the SimpleQuestions benchmark dataset and compare with baseline approaches.


\subsection{Data}
The SimpleQuestions~\cite{bordes2015large} dataset is released by the Facebook AI Research. It is the standard dataset on which almost all previous state-of-the-art literature reported their numbers~\cite{gupta2018retrieve,hao2018pattern}. It also represents the largest publicly available dataset for KBSQA with its size several orders of magnitude larger than other available datasets. It has $108,442$ simple questions with the corresponding facts from subsets of the Freebase (FB2M and FB5M). There are $1,837$ unique relations. We use the default train, validation and test partitions~\cite{bordes2015large} with $75,910$, $10,845$ and $21,687$ questions, respectively. We use FB2M with $2,150,604$ entities, $6,701$ relations and $14,180,937$ facts, respectively.

\subsection{Models}
For sequence tagging, we use the same BiLSTM-CRF model as the baseline~\cite{dai-li-xu:2016:P16-1} to label each word in the question as either subject or non-subject.
The configurations of the model (Table~\ref{tab:tagger_config}) basically follow the baseline~\cite{dai-li-xu:2016:P16-1}.
\begin{table}[t!]
\centering
\scalebox{0.86}{
\begin{tabular}{ll}
\toprule 
Vocab. size & 151,718 \\
Embedding dim & 300 \\
LSTM hidden dim & 256 \\
\# of LSTM layers & 2 \\
LSTM dropout & 0.5 \\
\# of CRF states & 4 (incl. start \& end) \\
\bottomrule
\end{tabular}
}
\caption{Sequence Tagger Configurations} \label{tab:tagger_config}
\end{table}

For subgraph selection, we use only unigrams of the tagged mention to retrieve the candidate facts (see Section~\ref{ssec:subgraph_rank}) and rank them by the proposed relevance score (\ref{eq:relevance_score}) with the tuned weight $\tau = 0.9$ (hence more emphasizing on literal matching). We select the facts with top-$n$ scores as the subgraphs and compare the corresponding recalls with the baseline method~\cite{yin-EtAl:2016:COLING}.

For fact selection, we employ a character-based CNN (CharCNN) model to score (mention, subject) pairs and a word-based CNN (WordCNN) model to score (pattern, relation) pairs (with model configurations shown in Table~\ref{tab:matcher_config}), which is similar to one of the state-of-the-art baselines AMPCNN~\cite{yin-EtAl:2016:COLING}. In fact, we first replicated the AMPCNN model and achieved comparable results, and then modified the AMPCNN model to take joint inputs and output scores directly (see Section~\ref{ssec:js_wellorder} and Figure~\ref{fig:js_model}).
Our CNN models have only two convolutional layers (versus six convolutional layers in the baseline) and have no attention mechanism, bearing much lower complexity than the baseline. The CharCNN and WordCNN differ only in the embedding layer, the former using character embeddings and the latter using word embeddings.
\begin{table}[t!]
\centering
\scalebox{0.86}{
\begin{tabular}{lll}
\toprule
Config. & CharCNN & WordCNN \\
\midrule
Alphabet / Vocab. size & 69 & 151,718 \\
Embedding dim & 60 & 300 \\
CNN layer 1 & (300, 3, 1, 1) & (1500, 3, 1, 1) \\
Activation & ReLU & ReLU \\
CNN layer 2 & (60, 3, 1, 1) & (300, 3, 1, 1) \\
AdaptiveMaxPool dim & 1 & 1 \\
\bottomrule
\end{tabular}
}
\caption{Matching Model Configurations} \label{tab:matcher_config}
\end{table}

The optimizer used for training the models is Adam~\cite{kingma2014adam}. The learning configurations are shown in Table~\ref{tab:learn_config}.
\begin{table}[t!]
\centering
\scalebox{0.86}{
\begin{tabular}{lll}
\toprule
Config. & Sequence Tagging & Matching \\
\midrule
Optimizer & Adam & Adam \\
Learning rate & 0.001 & 0.01 \\
Batch size & 64 & 32 \\
\# of epochs & 50 & 20 \\
\bottomrule
\end{tabular}
}
\caption{Learning Configurations} \label{tab:learn_config}
\end{table}

For the hyper-parameters shown in Table~\ref{tab:tagger_config}, \ref{tab:matcher_config} and \ref{tab:learn_config}, we basically follow the settings in baseline literature \cite{yin-EtAl:2016:COLING,dai-li-xu:2016:P16-1} to promote a fair comparison. Other hyper-parameters, such as the $\tau$ in the relevance score~(\ref{eq:relevance_score}), are tuned on the validation set.

Our proposed approach and the baseline approaches are evaluated in terms of (1) the top-$n$ subgraph selection recall (the percentage of questions that have the correct subjects in the top-$n$ candidates) and (2) the fact selection accuracy (i.e., the overall question answering accuracy).

\subsection{Results}
\textbf{Subgraph selection} The subgraph selection results for our approach and one of the state-of-the-art baselines~\cite{yin-EtAl:2016:COLING} are summarized in Table~\ref{tab:subgraph_select}.
Both the baseline and our approach use unigrams to retrieve candidates. The baseline ranks the candidates by the length of the longest common subsequence with heuristics while we rank the candidates by the joint relevance score defined in (\ref{eq:relevance_score}). We see that the literal score used in the baseline performs well and using the semantic score (the log-likelihood) (\ref{eq:prob_5}) only does not outperform the baseline (except for the top-$50$ case). This is due to the nature of how the questions in the SimpleQuestions dataset are generated---the majority of the questions only contain mentions matching the subjects in the Freebase in the lexical level, making the literal score sufficiently effective. However, we see that combining the literal score and semantic score outperforms the baseline by a large margin. For top-$1$, $5$, $10$, $20$, $50$ recall our ranking approach surpasses the baseline by $11.9$\%, $5.4$\%, $4.6$\%, $3.9$\%, $4.1$\%, respectively. Our approach also surpasses other baselines \cite{lukovnikov2017neural,yu2017improved,qu2018question,gupta2018retrieve} under the same settings.
We note that the recall is not monotonically increasing with the top-$n$. The reason is that, as opposed to conventional methods which rank the entire subgraph returned from unigram matching to select the top-$n$ candidates, we choose only the first $200$ candidates from the subgraph and then rank them with our proposed ranking score. This is more efficient, but at the price of potentially dropping the correct facts. One could trade efficiency for accuracy by ranking all the candidates in the subgraph.

\begin{table}[t!]
\centering
\scalebox{0.86}{
\begin{tabular}{lrl}
\toprule
Rank Method & Top-N & Recall \\
\midrule
\multirow{6}{*}{$|\sigma|$ + heuristics} & 1 & 0.736 \\
Literal: & 5 & 0.850 \\
& 10 & 0.874 \\
& 20 & 0.888 \\
\cite{yin-EtAl:2016:COLING} & 50 & 0.904 \\
& 100 & 0.916 \\
\midrule
\multirow{5}{*}{$\log{\mathbb{P}}$} & 1 & 0.482 \\
Semantic: & 10 & 0.753 \\
& 20 & 0.854 \\
& 50 & 0.921 \\
& 100 & 0.848\\
\midrule
\multirow{6}{*}{$0.9 |\sigma| + 0.1 \log{\mathbb{P}}$} & 1 & 0.855 \\
Joint: & 5 & 0.904 \\
& 10 & 0.920 \\
& 20 & 0.927 \\
& 50 & 0.945 \\
& 100 & 0.928 \\
\bottomrule
\end{tabular}
}
\caption{Subgraph Selection Results} \label{tab:subgraph_select}
\end{table}

\begin{table*}[t!]
\centering
\scalebox{0.86}{
\begin{tabular}{lllll}
\toprule
& Approach & Obj. & Sub. & Rel. \\
& & {\small(= {\bf Overall Acc.})} & & \\
\midrule
1 & AMPCNN & 76.4 & & \\
& \cite{yin-EtAl:2016:COLING} & & & \\
2 & BiLSTM & 78.1 & & \\
& \cite{petrochuk2018simplequestions} & & & \\
\midrule
3 & AMPCNN + wo-loss & 77.69 & & \\ 
4 & JS + wo-loss & 81.10 & 87.44 & 69.22 \\
5 & JS + wo-loss + sub50 & {\bf 85.44} & 91.47 & 76.98 \\
6 & JS + wo-loss + sub1 & 79.34 & 87.97 & 84.12 \\
\bottomrule
\end{tabular}
}
\caption{Fact Selection Accuracy (\%). The object accuracy is the end-to-end question answer accuracy, while subject and relation accuracies refer to separately computed subject accuracy and relation accuracy.} \label{tab:fact_select}
\end{table*}

\textbf{Fact selection} The fact selection results for our approach and baselines are shown in Table~\ref{tab:fact_select}. The object accuracy is the same as the overall question answer accuracy.
Recall that in Section~\ref{ssec:js_wellorder} we explained that the weight components in the well-order loss~(\ref{eq:objective_wellorder}) are adjusted dynamically in the training to impose a larger penalty for mention--subject mismatches and hence enforce correct matches. This can be observed by looking at the different loss components and weights as well the subject and relation matching accuracies during the training. As weights for mention--subject matches increase, the losses for mention--subject matches also increase, while both the errors for mention--subject matches and pattern--relation matches are high. To reduce the errors, the model is forced to match mention--subject more correctly. As a result, the corresponding weights and losses decrease, and both mention--subject and pattern--relation match accuracies increase.

\textbf{\textit{Effectiveness of well-order loss and joint-scoring model}}  The first and second row of Table~\ref{tab:fact_select} are taken from the baseline AMPCNN~\cite{yin-EtAl:2016:COLING} and BiLSTM~\cite{petrochuk2018simplequestions} (the state of the art prior to our work\footnote{As noted, \citet{ture-jojic:2017:EMNLP2017} reported better performance than us but neither \citet{petrochuk2018simplequestions} nor \citet{mohammed2017strong} could replicate their result.}). The third row shows the accuracy of the baseline with our proposed well-order loss and we see a $1.3$\% improvement, demonstrating the effectiveness of the well-order loss.
Further, the fourth row shows the accuracy of our joint-scoring (JS) model with well-order loss and we see a $3$\% improvement over the best baseline\footnote{At the time of submission we also found that \citet{hao2018pattern} reported 80.2\% accuracy.}, demonstrating the effectiveness of the joint-scoring model. 

\textbf{\textit{Effectiveness of subgraph ranking}} The fifth row of Table~\ref{tab:fact_select} shows the accuracy of our joint-scoring model with well-order loss and top-$50$ ranked subgraph and we see a further $4.3$\% improvement over our model without subgraph ranking (the fourth row), and a $7.3$\% improvement over the best baseline. In addition, the subject accuracy increases by $4.0$\%, which is due to the subgraph ranking. Interestingly, the relation accuracy increases by $7.8$\%, which supports our claim that improving subject matching can improve relation matching. This demonstrates the effectiveness of our subgraph ranking and joint-scoring approach. The sixth row shows the accuracy of our joint-scoring model with well-order loss and only the top-$1$ subject. In this case, the subject accuracy is limited by the top-$1$ recall which is $85.5$\%. Despite that, our approach outperforms the best baseline by $1.2$\%. Further, the relation accuracy increases by $7.1$\% over the fifth row, because restricting the subject substantially confines the choice of relations. This shows that a sufficiently high top-$1$ subgraph recall reduces the need for subject matching.


\subsection{Error Analysis}
In order to analyze what constitutes the errors of our approach, we select the questions in the test set for which our best model has predicted wrong answers, and analyze the source of errors (see Table~\ref{tab:err_decomp}).
We observe that the errors can be categorized as follows: (1) Incorrect subject prediction; however, some subjects are actually correct, e.g., the prediction ``New York'' v.s.\@ ``New York City.'' (2) Incorrect relation prediction; however, some relations are actually correct, e.g., the prediction ``fictional-universe.fictional-character.character-created-by'' v.s.\@ ``book.written-work.author'' in the question ``Who was the writer of Dark Sun?'' and ``music.album.genre'' v.s.\@ ``music.artist.genre.'' (3) Incorrect prediction of both.

\begin{table}[t!]
\centering
\scalebox{0.86}{
\begin{tabular}{lr}
\toprule 
 Incorrect Sub. only & 8.67 \\
 Incorrect Rel. only & 16.26 \\
 Incorrect Sub. \& Rel. & 34.50 \\
 Other & 40.57 \\
\bottomrule
\end{tabular}
}
\caption{Error Decomposition (\%). Percentages for total of $3157$ errors.} \label{tab:err_decomp}
\end{table}

However, these three reasons only make up 59.43\% of the errors. The other 40.57\% errors are due to:
(4) Ambiguous questions, which take up the majority of the errors, e.g., ``Name a species of fish.'' or ``What movie is a short film?'' These questions are too general and can have multiple correct answers. Such issues in the SimpleQuestions dataset are analyzed by \citet{petrochuk2018simplequestions} (see further discussion on this at the end of this Section).
(5) Non-simple questions, e.g., ``Which drama film was released in 1922?'' This question requires two KB facts instead of one to answer correctly.
(6) Wrong fact questions where the reference fact is non-relevant, e.g., ``What is an active ingredient in Pacific?'' is labeled with ``Triclosan 0.15 soap''. 
(7) Out of scope questions, which have entities or relations out the scope of FB2M.
(8) Spelling inconsistencies, e.g., the predicted answer ``Operation Shylock: A Confession'' v.s.\@ the reference answer ``Operation Shylock'', and the predicted answer ``Tom and Jerry: Robin Hood and His Merry Mouse'' v.s.\@ the reference answer ``Tom and Jerry''.
For these cases, even when the models predict the subjects and relations correctly, these questions are fundamentally \emph{unanswerable}.

Although these issues are inherited from the dataset itself, given the large size of the dataset and the small proportion of the problematic questions, it is sufficient to validate the reliability and significance of our performance improvement and conclusions.

\textbf{Answerable Questions Redefined} \citet{petrochuk2018simplequestions} set an upper bound of 83.4\% for the accuracy on the SimpleQuestions dataset. However, our models are able to do better than the upper bound. Are we doing something wrong? \citet{petrochuk2018simplequestions} claim that a question is unanswerable if there exist multiple valid subject--relation pairs in the knowledge graph, but we claim that a question is unanswerable if and only if there is no valid fact in the knowledge graph. There is a subtle difference between these two claims.

Based on different definitions of answerable questions, we further claim that incorrect subject or incorrect relation can still lead to a correct answer. For example, for the question ``What is a song from Hier Komt De Storm?'' with the fact (Hier Komt De Storm: 1980-1990 live, music.release.track-list, Stephanie), our predicted subject ``Hier Komt De Storm: 1980-1990 live'' does not match the reference subject ``Hier Komt De Storm'', but our model predicts the correct answer ``Stephanie'' because it can deal with inexact match of the subjects.
In the second example, for the question ``Arkham House is the publisher behind what novel?'', our predicted relation ``book.book-edition.publisher'' does not match the reference relation ``book.publishing-company.books-published'', but our model predicts the correct answer ``Watchers at the Strait Gate'' because it can deal with paraphrases of relations.
In the third example, for the question ``Who was the king of Lydia and Croesus's father?'', the correct subject ``Croesus'' ranks second in our subject predictions and the correct relation ``people.person.parents'' ranks fourth in our relation predictions, but our model predicts the correct answer ``Alyattes of Lydia'' because it reweighs the scores with respect to the subject--relation dependency and the combined score of subject and relation ranks first.

To summarize, the reason that we are able to redefine answerable questions and achieve significant performance gain is that we take advantage of the subgraph ranking and the subject--relation dependency.

\section{Conclusions}
In this work, we propose a subgraph ranking method and joint-scoring approach to improve the performance of KBSQA.
The ranking method combines literal and semantic scores to deal with inexact match and achieves better subgraph selection results than the state of the art.
The joint-scoring model with well-order loss couples the dependency of subject matching and relation matching and enforces the order of scores.
Our proposed approach achieves a new state of the art on the SimpleQuestions dataset, surpassing the best baseline by a large margin.

In the future work, one could further improve the performance on simple question answering tasks by exploring relation ranking, different embedding strategies and network structures, dealing with open questions and out-of-scope questions.
One could also consider extending our approach to complex questions, e.g., multi-hop questions where more than one supporting facts is required. Potential directions may include ranking the subgraph by assigning each edge (relation) a closeness score and evaluating the length of the shortest path between any two path-connected entity nodes.

\section*{Acknowledgments}
The authors would like to thank anonymous reviewers. The authors would also like to thank Nikko Ström and other Alexa AI team members for their feedback.

\bibliography{main}

\begin{thebibliography}{31}
\expandafter\ifx\csname natexlab\endcsname\relax\def\natexlab#1{#1}\fi

\bibitem[{Berant et~al.(2013)Berant, Chou, Frostig, and
  Liang}]{berant2013semantic}
Jonathan Berant, Andrew Chou, Roy Frostig, and Percy Liang. 2013.
\newblock Semantic parsing on {Freebase} from question-answer pairs.
\newblock In \emph{Proceedings of the 2013 Conference on Empirical Methods in
  Natural Language Processing}, pages 1533--1544.

\bibitem[{Bordes et~al.(2014)Bordes, Chopra, and
  Weston}]{bordes-chopra-weston:2014:EMNLP2014}
Antoine Bordes, Sumit Chopra, and Jason Weston. 2014.
\newblock \href {http://www.aclweb.org/anthology/D14-1067} {Question answering
  with subgraph embeddings}.
\newblock In \emph{Proceedings of the 2014 Conference on Empirical Methods in
  Natural Language Processing}, pages 615--620, Doha, Qatar. Association for
  Computational Linguistics.

\bibitem[{Bordes et~al.(2015)Bordes, Usunier, Chopra, and
  Weston}]{bordes2015large}
Antoine Bordes, Nicolas Usunier, Sumit Chopra, and Jason Weston. 2015.
\newblock Large-scale simple question answering with memory networks.
\newblock \emph{arXiv preprint arXiv:1506.02075}.

\bibitem[{Cao et~al.(2006)Cao, Xu, Liu, Li, Huang, and Hon}]{cao2006adapting}
Yunbo Cao, Jun Xu, Tie-Yan Liu, Hang Li, Yalou Huang, and Hsiao-Wuen Hon. 2006.
\newblock Adapting ranking {SVM} to document retrieval.
\newblock In \emph{Proceedings of the 29th Annual International ACM SIGIR
  Conference on Research and Development in Information Retrieval}, pages
  186--193. ACM.

\bibitem[{Collobert and Weston(2008)}]{collobert2008unified}
Ronan Collobert and Jason Weston. 2008.
\newblock A unified architecture for natural language processing: Deep neural
  networks with multitask learning.
\newblock In \emph{Proceedings of the 25th International Conference on Machine
  Learning}, pages 160--167. ACM.

\bibitem[{Dai et~al.(2016)Dai, Li, and Xu}]{dai-li-xu:2016:P16-1}
Zihang Dai, Lei Li, and Wei Xu. 2016.
\newblock \href {http://www.aclweb.org/anthology/P16-1076} {{CFO}: Conditional
  focused neural question answering with large-scale knowledge bases}.
\newblock In \emph{Proceedings of the 54th Annual Meeting of the Association
  for Computational Linguistics}, volume 1: Long Papers, pages 800--810,
  Berlin, Germany. Association for Computational Linguistics.

\bibitem[{Dong et~al.(2015)Dong, Wei, Zhou, and Xu}]{dong2015question}
Li~Dong, Furu Wei, Ming Zhou, and Ke~Xu. 2015.
\newblock Question answering over {Freebase} with multi-column convolutional
  neural networks.
\newblock In \emph{Proceedings of the 53rd Annual Meeting of the Association
  for Computational Linguistics and the 7th International Joint Conference on
  Natural Language Processing}, volume 1: Long Papers, pages 260--269.

\bibitem[{Gupta et~al.(2018)Gupta, Chinnakotla, and
  Shrivastava}]{gupta2018retrieve}
Vishal Gupta, Manoj Chinnakotla, and Manish Shrivastava. 2018.
\newblock Retrieve and re-rank: A simple and effective ir approach to simple
  question answering over knowledge graphs.
\newblock In \emph{Proceedings of the First Workshop on Fact Extraction and
  Verification (FEVER)}, pages 22--27.

\bibitem[{Hao et~al.(2018)Hao, Liu, He, Liu, and Zhao}]{hao2018pattern}
Yanchao Hao, Hao Liu, Shizhu He, Kang Liu, and Jun Zhao. 2018.
\newblock Pattern-revising enhanced simple question answering over knowledge
  bases.
\newblock In \emph{Proceedings of the 27th International Conference on
  Computational Linguistics}, pages 3272--3282.

\bibitem[{Hao et~al.(2017)Hao, Zhang, Liu, He, Liu, Wu, and Zhao}]{hao2017end}
Yanchao Hao, Yuanzhe Zhang, Kang Liu, Shizhu He, Zhanyi Liu, Hua Wu, and Jun
  Zhao. 2017.
\newblock An end-to-end model for question answering over knowledge base with
  cross-attention combining global knowledge.
\newblock In \emph{Proceedings of the 55th Annual Meeting of the Association
  for Computational Linguistics}, volume 1: Long Papers, pages 221--231.

\bibitem[{He and Golub(2016)}]{he2016character}
Xiaodong He and David Golub. 2016.
\newblock Character-level question answering with attention.
\newblock In \emph{Proceedings of the 2016 Conference on Empirical Methods in
  Natural Language Processing}, pages 1598--1607.

\bibitem[{Hu et~al.(2018)Hu, Zou, Yu, Wang, and Zhao}]{hu2018answering}
Sen Hu, Lei Zou, Jeffrey~Xu Yu, Haixun Wang, and Dongyan Zhao. 2018.
\newblock Answering natural language questions by subgraph matching over
  knowledge graphs.
\newblock \emph{IEEE Transactions on Knowledge and Data Engineering},
  30(5):824--837.

\bibitem[{Huang et~al.(2015)Huang, Xu, and Yu}]{huang2015bidirectional}
Zhiheng Huang, Wei Xu, and Kai Yu. 2015.
\newblock Bidirectional {LSTM-CRF} models for sequence tagging.
\newblock \emph{arXiv preprint arXiv:1508.01991}.

\bibitem[{Khashabi et~al.(2016)Khashabi, Khot, Sabharwal, Clark, Etzioni, and
  Roth}]{khashabi2016question}
Daniel Khashabi, Tushar Khot, Ashish Sabharwal, Peter Clark, Oren Etzioni, and
  Dan Roth. 2016.
\newblock Question answering via integer programming over semi-structured
  knowledge.
\newblock In \emph{Proceedings of the Twenty-Fifth International Joint
  Conference on Artificial Intelligence}, pages 1145--1152. AAAI Press.

\bibitem[{Kingma and Ba(2014)}]{kingma2014adam}
Diederik~P Kingma and Jimmy Ba. 2014.
\newblock Adam: A method for stochastic optimization.
\newblock \emph{arXiv preprint arXiv:1412.6980}.

\bibitem[{Kumar et~al.(2016)Kumar, Irsoy, Ondruska, Iyyer, Bradbury, Gulrajani,
  Zhong, Paulus, and Socher}]{kumar2016ask}
Ankit Kumar, Ozan Irsoy, Peter Ondruska, Mohit Iyyer, James Bradbury, Ishaan
  Gulrajani, Victor Zhong, Romain Paulus, and Richard Socher. 2016.
\newblock Ask me anything: Dynamic memory networks for natural language
  processing.
\newblock In \emph{International Conference on Machine Learning}, pages
  1378--1387.

\bibitem[{Lukovnikov et~al.(2017)Lukovnikov, Fischer, Lehmann, and
  Auer}]{lukovnikov2017neural}
Denis Lukovnikov, Asja Fischer, Jens Lehmann, and S{\"o}ren Auer. 2017.
\newblock Neural network-based question answering over knowledge graphs on word
  and character level.
\newblock In \emph{Proceedings of the 26th International Conference on World
  Wide Web}, pages 1211--1220. International World Wide Web Conferences
  Steering Committee.

\bibitem[{Mohammed et~al.(2018)Mohammed, Shi, and Lin}]{mohammed2017strong}
Salman Mohammed, Peng Shi, and Jimmy Lin. 2018.
\newblock Strong baselines for simple question answering over knowledge graphs
  with and without neural networks.
\newblock In \emph{Proceedings of the 2018 Conference of the North American
  Chapter of the Association for Computational Linguistics: Human Language
  Technologies}, volume 2: Short Papers, pages 291--296.

\bibitem[{Pennington et~al.(2014)Pennington, Socher, and
  Manning}]{pennington-socher-manning:2014:EMNLP2014}
Jeffrey Pennington, Richard Socher, and Christopher Manning. 2014.
\newblock \href {http://www.aclweb.org/anthology/D14-1162} {{GloVe}: Global
  vectors for word representation}.
\newblock In \emph{Proceedings of the 2014 Conference on Empirical Methods in
  Natural Language Processing}, pages 1532--1543, Doha, Qatar. Association for
  Computational Linguistics.

\bibitem[{Petrochuk and Zettlemoyer(2018)}]{petrochuk2018simplequestions}
Michael Petrochuk and Luke Zettlemoyer. 2018.
\newblock {SimpleQuestions} nearly solved: A new upperbound and baseline
  approach.
\newblock \emph{arXiv preprint arXiv:1804.08798}.

\bibitem[{Qu et~al.(2018)Qu, Liu, Kang, Shi, and Ye}]{qu2018question}
Yingqi Qu, Jie Liu, Liangyi Kang, Qinfeng Shi, and Dan Ye. 2018.
\newblock Question answering over {Freebase} via attentive {RNN} with
  similarity matrix based {CNN}.
\newblock \emph{arXiv preprint arXiv:1804.03317}.

\bibitem[{Sukhbaatar et~al.(2015)Sukhbaatar, Weston, Fergus
  et~al.}]{sukhbaatar2015end}
Sainbayar Sukhbaatar, Jason Weston, Rob Fergus, et~al. 2015.
\newblock End-to-end memory networks.
\newblock In \emph{Advances in Neural Information Processing Systems}, pages
  2440--2448.

\bibitem[{Ture and Jojic(2017)}]{ture-jojic:2017:EMNLP2017}
Ferhan Ture and Oliver Jojic. 2017.
\newblock \href {https://www.aclweb.org/anthology/D17-1307} {No need to pay
  attention: Simple recurrent neural networks work!}
\newblock In \emph{Proceedings of the 2017 Conference on Empirical Methods in
  Natural Language Processing}, pages 2866--2872, Copenhagen, Denmark.
  Association for Computational Linguistics.

\bibitem[{Vu et~al.(2016)Vu, Gupta, Adel, and Sch{\"u}tze}]{vu2016bi}
Ngoc~Thang Vu, Pankaj Gupta, Heike Adel, and Hinrich Sch{\"u}tze. 2016.
\newblock Bi-directional recurrent neural network with ranking loss for spoken
  language understanding.
\newblock In \emph{2016 IEEE International Conference on Acoustics, Speech and
  Signal Processing (ICASSP)}, pages 6060--6064. IEEE.

\bibitem[{Yao and Van~Durme(2014)}]{yao2014information}
Xuchen Yao and Benjamin Van~Durme. 2014.
\newblock Information extraction over structured data: Question answering with
  {Freebase}.
\newblock In \emph{Proceedings of the 52nd Annual Meeting of the Association
  for Computational Linguistics}, volume 1: Long Papers, pages 956--966.

\bibitem[{Yih et~al.(2015)Yih, Chang, He, and Gao}]{yih2015semantic}
Wentau Yih, Minwei Chang, Xiaodong He, and Jianfeng Gao. 2015.
\newblock Semantic parsing via staged query graph generation: Question
  answering with knowledge base.
\newblock In \emph{Proceedings of the 53rd Annual Meeting of the Association
  for Computational Linguistics and the 7th International Joint Conference on
  Natural Language Processing}, volume 1: Long Papers, pages 1321--1331.

\bibitem[{Yin et~al.(2016)Yin, Yu, Xiang, Zhou, and
  Sch\"{u}tze}]{yin-EtAl:2016:COLING}
Wenpeng Yin, Mo~Yu, Bing Xiang, Bowen Zhou, and Hinrich Sch\"{u}tze. 2016.
\newblock \href {http://aclweb.org/anthology/C16-1164} {Simple question
  answering by attentive convolutional neural network}.
\newblock In \emph{Proceedings of COLING 2016, the 26th International
  Conference on Computational Linguistics: Technical Papers}, pages 1746--1756,
  Osaka, Japan. The COLING 2016 Organizing Committee.

\bibitem[{Yu et~al.(2017)Yu, Yin, Hasan, dos Santos, Xiang, and
  Zhou}]{yu2017improved}
Mo~Yu, Wenpeng Yin, Kazi~Saidul Hasan, Cicero dos Santos, Bing Xiang, and Bowen
  Zhou. 2017.
\newblock Improved neural relation detection for knowledge base question
  answering.
\newblock In \emph{Proceedings of the 55th Annual Meeting of the Association
  for Computational Linguistics}, volume 1: Long Papers, pages 571--581.

\bibitem[{Zhang et~al.(2018)Zhang, Dai, Kozareva, Smola, and
  Song}]{zhang2018variational}
Yuyu Zhang, Hanjun Dai, Zornitsa Kozareva, Alexander~J Smola, and Le~Song.
  2018.
\newblock Variational reasoning for question answering with knowledge graph.
\newblock In \emph{Thirty-Second AAAI Conference on Artificial Intelligence}.

\bibitem[{Zhao et~al.(2015)Zhao, Huang, Wang, and Tan}]{zhao2015deep}
Fang Zhao, Yongzhen Huang, Liang Wang, and Tieniu Tan. 2015.
\newblock Deep semantic ranking based hashing for multi-label image retrieval.
\newblock In \emph{Proceedings of the IEEE Conference on Computer Vision and
  Pattern Recognition}, pages 1556--1564.

\bibitem[{Zheng et~al.(2018)Zheng, Yu, Zou, and Cheng}]{zheng2018question}
Weiguo Zheng, Jeffrey~Xu Yu, Lei Zou, and Hong Cheng. 2018.
\newblock Question answering over knowledge graphs: Question understanding via
  template decomposition.
\newblock \emph{Proceedings of the VLDB Endowment}, 11(11).

\end{thebibliography}
\bibliographystyle{acl_natbib}

\end{document}